\pdfoutput=1

\documentclass[11pt]{article}

\usepackage{emnlp2022}

\usepackage{times}
\usepackage{latexsym}

\usepackage{amsmath}

\usepackage{times}
\usepackage{latexsym}
\usepackage{enumitem}
\usepackage[T1]{fontenc}
\usepackage[utf8]{inputenc}

\usepackage{xspace}
\usepackage{todonotes}
\usepackage{booktabs}
\usepackage{soul}
\usepackage{algorithm}
\usepackage{algorithmic}
\usepackage{newfloat}
\usepackage{listings}

\usepackage{booktabs}
\usepackage{graphicx}
\usepackage{subcaption} 
\usepackage{fontawesome} 
\usepackage{comment}

\newcommand\eat[1]{}

\usepackage[T1]{fontenc}

\usepackage[utf8]{inputenc}

\usepackage{microtype}

\title{GREENER: Graph Neural Networks for News Media Profiling}

\author{Panayot Panayotov \\
  Sofia University \\
  \texttt{p.d.panayotov09@gmail.com} \\\And
  Utsav Shukla \\
  TIET \\
   \\ \And
  Husrev T. Sencar \\
  QCRI, HBKU\\
  \texttt{hsencar@hbku.edu.qa} \\\AND
   Mohamed Nabeel  \\
  QCRI, HBKU \\
  \texttt{mnabeel@hbku.edu.qa} \\\And
      Preslav Nakov \\
  Mohamed bin Zayed University of Artificial Intelligence \\
  \texttt{preslav.nakov@mbzuai.ac.ae} \\}

\begin{document}
\maketitle
\vspace{10cm}
\begin{abstract}
We study the problem of profiling news media on the Web with respect to their factuality of reporting and bias. This is an important but under-studied problem related to disinformation and ``fake news'' detection, but it addresses the issue at a coarser granularity compared to looking at an individual article or an individual claim. This is useful as it allows to profile entire media outlets in advance. Unlike previous work, which has focused primarily on text (e.g.,~on the text of the articles published by the target website, or on the textual description in their social media profiles or in Wikipedia), here our main focus is on modeling the similarity between media outlets based on the overlap of their audience. This is motivated by homophily considerations, i.e.,~the tendency of people to have connections to people with similar interests, which we extend to media, hypothesizing that similar types of media would be read by similar kinds of users. In particular, we propose GREENER (GRaph nEural nEtwork for News mEdia pRofiling), a model that builds a graph of inter-media connections based on their audience overlap, and then uses graph neural networks to represent each medium. 
We find that such representations are quite useful for predicting the factuality and the bias of news media outlets, yielding improvements over state-of-the-art results reported on two datasets.  
When augmented with conventionally used representations obtained from news articles, Twitter, YouTube, Facebook, and Wikipedia, prediction accuracy is found to improve by 2.5-27 macro-F1 points for the two tasks.

\end{abstract}

\section{Introduction}

The problem of news media profiling with respect to their factuality of reporting and political bias is important but under-studied. It is related to disinformation and ``fake news'' detection, but it is of different granularity compared to looking at an individual article or at an individual claim. 
This kind of profiling can be done by professional fact-checkers, who inspect the articles and the multimedia material published by the target news outlet. 

However, doing this automatically while solely relying on text features is a very challenging task as previous work has shown \cite{Ramy2018,baly2020written}. It gets even more complicated when considering news sources where only limited amount of content is available for evaluation.
Therefore, not only is there a need to more thoroughly characterize news media, but there is also a need to be able to do so in a predictive fashion using limited information.

A crucial consideration 
is the need to complement the textual representation with other elements of a news medium that may serve as reliable indicators of its factuality of reporting and bias. These may relate to multimedia creation and curation processes \cite{jin2016novel, Huh_2018_ECCV}, to its underlying infrastructure and technological components used to serve its content \cite{fairbanks2018credibility,castelo2019topic,hounsel2020identifying}, and, more critically, to characteristics of its audience \cite{baly2020written,chen2020proactive}. 

Here, we explore ways to augment the textual representations from the articles published by a target news medium by introducing new information sources that relate to media audience homophily, audience engagement, and media popularity. In particular, we propose the GREENER (GRaph nEural nEtwork for News mEdia pRofiling) model, which builds graph neural networks that model the audience overlap between websites, which we further complement with other state-of-the-art representations.
Our contributions are as follows:

\begin{itemize}
    \item We propose a novel model, based on graph neural networks that models the audience overlap between media in order to predict the factuality and the bias of entire news outlets.
    \item We show that the information in our graph is complementary to other information sources such as the text of the articles by the target news outlet, as well as to information from Twitter, Youtube, Facebook, and Wikipedia.
    \item We report sizable improvements over the state of the art on two standard datasets and for two tasks: predicting the factuality of reporting and the bias of news outlets.
    \item We release the code, the data, the processed features, and the representations used in our experiments\eat{\textcolor{red}{(https://anonymous/)}}.
\end{itemize}


\section{Related Work}
\label{sec:related_work}

Previous work on automating the process of characterizing news sites based on the factuality of their reporting and on their political bias has mainly focused 
on analysis of the textual content of the respective website~\cite{afroz2012detecting, rubin2015towards,rashkin2017truth, potthast2018stylometric, Ramy2018, perez2019automatic}.
Although style-based analysis of the text can help reveal the intent of an article, it cannot ultimately evaluate the authenticity and the objectivity of the claims stated in that article. 
In fact, as demonstrated by the results in \cite{baly2020written} on a manually fact-checked and categorized dataset, state-of-the-art textual representations can only achieve a prediction accuracy around 65-71\% for factuality and 70-85\% for bias depending on the datasets. 
Thus, several approaches have been proposed to supplement the content-level analysis with other contextual and relational information available about the target news outlet. 

Multimedia has been an important element of conveying news and information by all news media.
Due to its prevalence, tampering detection and identification of processing related traces in photos and videos have long been a focus of study \cite{dif}.
The fact that multimedia editors of a news site follow a workflow when creating, acquiring, editing, and curating content for their pages makes it possible to
characterize a website based on multimedia content. 
Therefore, visual features are increasingly being explored and used to predict the factuality of reporting of news media
\cite{jin2016novel,Huh_2018_ECCV,khattar2019mvae,zlatkova2019fact,qi2019exploiting,singhal2019spotfake}.  

Beyond textual and visual features, news sites also exhibit distinct characteristics in the way they set up their infrastructure to serve content.
To detect low-factuality news sites, it was proposed to use features that relate to network, web design, and data elements of the target website.
At the network level, it was shown that a website's domain, certificate, and hosting properties can serve as reliable identifiers \cite{hounsel2020identifying}.
Concerning the web design aspect, several features capturing the pattern of elements that govern the structure and the style of a web page have been also used \cite{castelo2019topic}.
Finally, at the data level, shared content among web sites and mutually linked sites were used to identify similar sites \cite{fairbanks2018credibility}.
Overall, a major advantage of using infrastructure features is their content-agnostic nature. 

Another set of features used to estimate the factuality and the bias of a news source is based on audience characteristics following the homophily principle, which simply states that similar individuals interact with each other at a higher rate than with dissimilar ones. 
In the context of social media platforms, several approaches were proposed to infer the similarity between news media through obtaining and comparing descriptive characteristics of the followers of a news medium \cite{baly2020written} and by profiling how these followers respond to the content of the target news medium in their comments and with their posting and sharing behavior \cite{wong2013quantifying, chen2020proactive}.
In this regard, a more reliable indicator for similarity of news sites is how much the followers of different news media overlap \cite{darwish2020unsupervised}.

Ultimately, these features were all obtained from disparate data sources and are all complementary in nature. 
Therefore, a more accurate characterization of the news reporting practice of a given news medium can be achieved by deploying more comprehensive heterogeneous learning approaches. 
To this objective, in this work, we propose to use graph neural networks to model the audience homophily relations based on audience overlap and engagement statistics from Alexa. 
In order to provide a more holistic view, our representation is also coupled with state-of-the-art textual representations extracted from media articles, as well as on other audience characteristics proposed in the context of social media platforms.

\section{Method}
\label{sec:method}

To characterize the similarity between news media in terms of their factuality of reporting and political bias, we mainly rely on audience overlap, 
which is based on the idea that if a group of visitors have a common interest in some websites, then those websites must be similar in some respect. 
With this idea, we create an undirected Web audience overlap graph, where nodes represent news media sites and edges indicate that 
that two news sites have an overlapping set of visitors, as well as the degree of overlap.
The graph is created using a seed list of news sites for which factuality and bias ratings are manually annotated by professional fact-checkers\footnote{The annotation of the seed nodes in the graph comes from factuality and the bias ratings provided at \url{http://mediabiasfactcheck.com}}.
This initial graph only captures the relation between websites due to visitors that are interested in a pair of sites, and it cannot represent indirect relations where visitors might have common taste in their news consumption, but do not necessarily visit the same websites.

In order to also identify such connections between news sites, we iteratively expand the graph by adding new neighboring nodes for a more comprehensive representation of the audience overlap, which is discussed in detail in section 3.2. 
The graph is further enhanced by incorporating user engagement statistics as node attributes in order to model the relation between a site and its visitors better.
We then use graph neural networks to encode these relations and to obtain node embeddings representing different categories of news sites.
We further combine these embeddings with textual representations 
from articles from each news website. 




\subsection{Alexa Metrics}
\begin{figure}[h]
\centering
\includegraphics[scale=0.08]{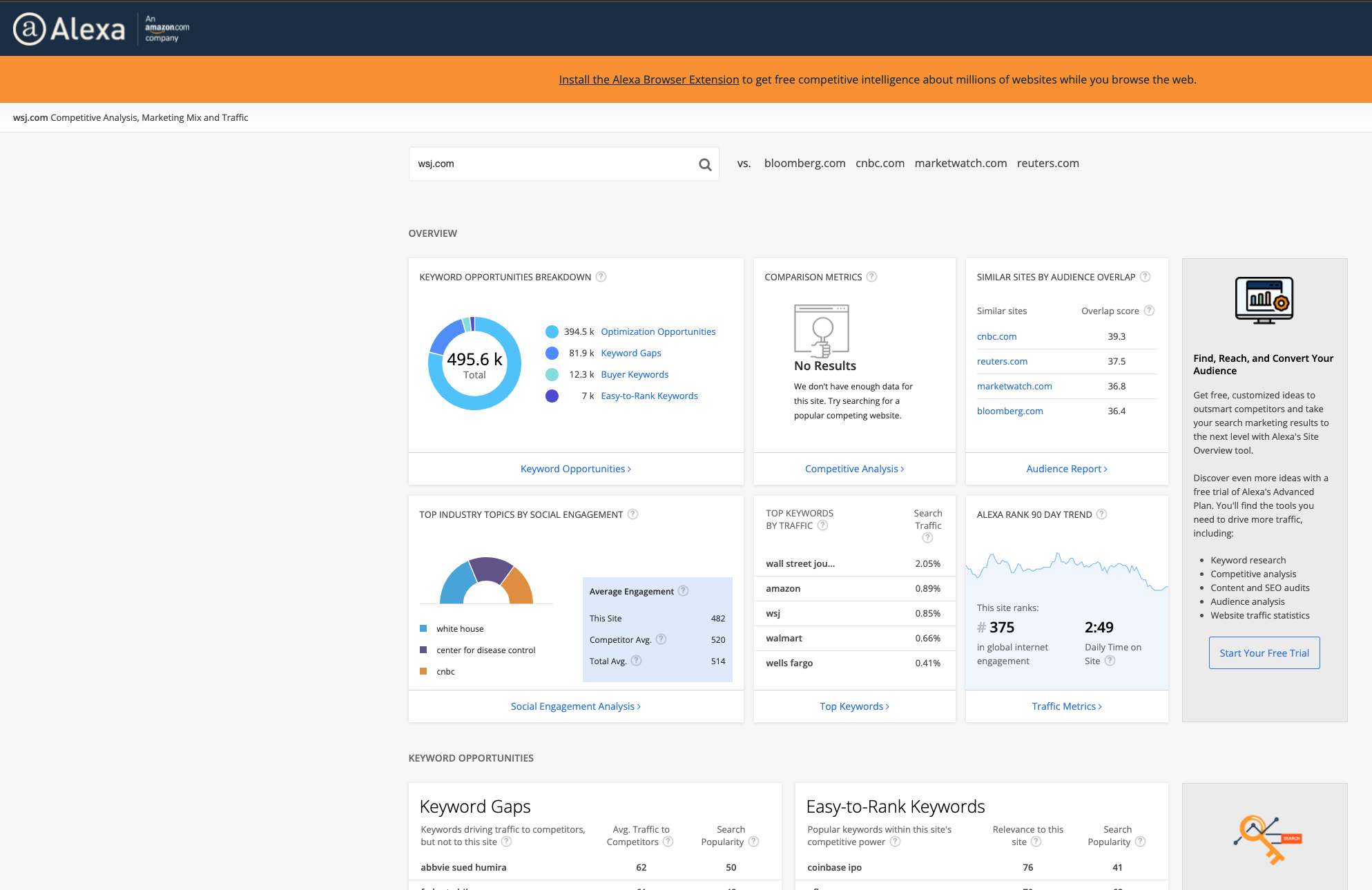}
\caption{Alexa Rank information for \emph{wsj.com}.}
\label{fig:alexa-info-page}
\end{figure}

Alexa is a web traffic analysis company that produces statistics about the browsing behavior of Internet users. These statistics are computed over a rolling three-month window; they are updated daily, and are either obtained directly from sites that choose to install a tracking script on their web pages or are estimated from a sample of data generated by millions of users using 
browser extensions and plug-ins related to Alexa.\footnote{\url{www.alexa.com/find-similar-sites}} 
Figure~\ref{fig:alexa-info-page} shows a sample Alexa page providing web traffic and domain statistics for the website wsj.com. 

We used the Alexa Audience Overlap Tool to extract statistics, which we used to build our Web audience overlap graph: links and node attributes.

{\em  Audience Overlap:} 
This includes a list of websites 
most similar to the target. Alexa calculates the similarity between two websites based on shared visitors and overlap in the keywords used in their webpages. For each pair of overlapping sites, a score is computed to quantify the degree of overlap. 
Preliminary analysis of Alexa Rank has shown that a highly factual site, such as reuters.com, has sizable audience overlap with other factual sites. Similarly, a low-factuality website such as \emph{infowars.com}, shares audience with other low-factulity websites. The audience homophily also holds for political bias, e.g., \emph{foxnews.com} and \emph{cnn.com} share audience primarily with other right- and left-leaning websites, respectively.

\begin{figure}[h]
\centering
\includegraphics[scale=0.12]{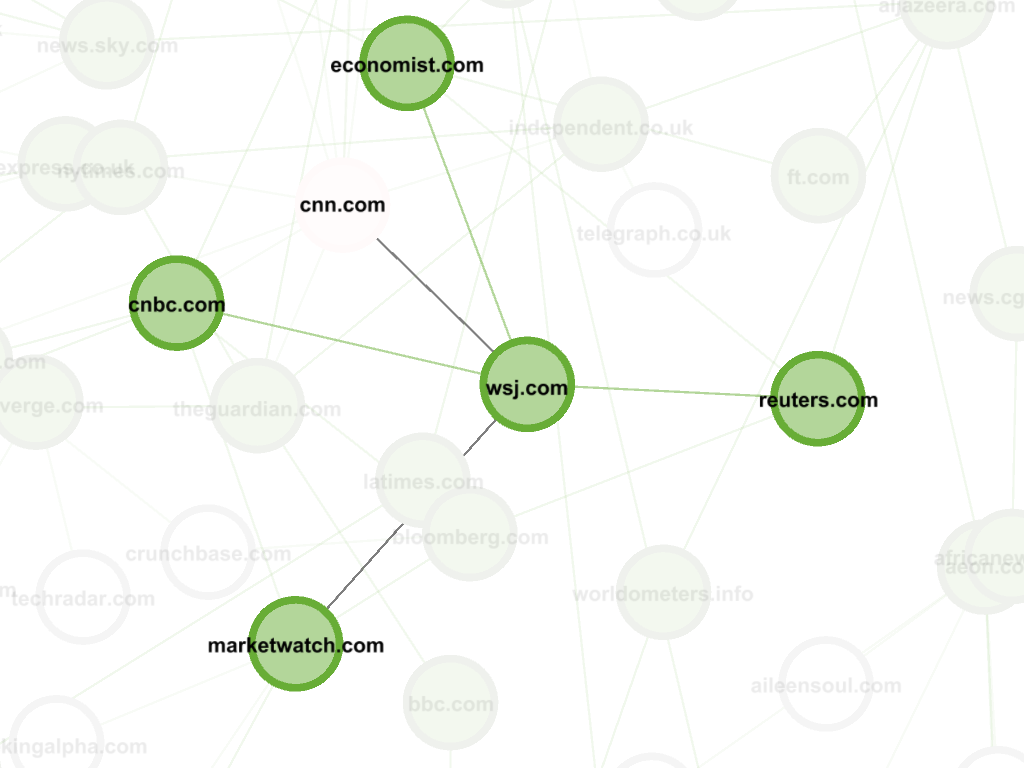}
\caption{Audience overlap graph for \emph{The Wall Street Journal}, showing that most of its neighboring nodes have the same factuality label: \emph{high}.}
\label{fig:wsj}
\end{figure}

Figure~\ref{fig:wsj} shows the overlapping websites for wsj.com, where we can see its homophily with other high-factuality websites. A similar pattern is observed for bias, where left/right-leaning websites 
overlap with other left/right-leaning websites.

{\em Traffic Rank:}
A site's rank is a measure of its popularity, which is computed based on the number of unique users that visit it and on the total number of URL requests they made on a single day. Page views corresponding to different URL requests are counted separately only if they are 30 minutes apart from each other. We logarithmically scale this rank for a more compact representation.

{\em Sites Linking In:}
This is the number of websites in the Common Crawl corpus that link to a given website. 
The list excludes links placed to influence search engine rankings of the linked page.

{\em Bounce Rate:}
Bounce rate is an engagement statistic showing the level of interest visitors have in the content of a website.
It is measured as the percentage of visits that consist of a single pageview, {\em i.e.},~when the visitor does not click on any of the links on the landing page.

{\em Daily Pageviews per Visitor:} 
This is the average number of pages viewed (or refreshed) by visitors.

{\em Daily Time on Site:} 
This is another engagement statistics, which shows the average time, in minutes and seconds, that a visitor spends on a target website each day. We convert it to seconds.


{\em Binarized Alexa Metrics:} 
Among the above-described Alexa site metrics, \emph{Sites Linking In} produces a list of websites through analysis of web crawled data. 
Therefore, the completeness of the list depends on the crawling coverage.
The last three metrics, ({\em i.e.},~daily page views, bounce rate, and daily time on site) 
measure the level of user engagement with the 
website. 
If users bounce at a higher rate, do not stay very long, or only view a few pages, 
they are likely less interested in that website.
Hence, the reliability of these three metrics depends on the size of the sample of users that was used for the measurements.
Due to these limitations, not all sites have such corresponding metrics calculated by Alexa Rank.
Table~\ref{table:nonlin} shows statistics about the overall availability of these metrics for websites in the two datasets. 
Therefore, as a more crude measure of site popularity and engagement, we also use the binary versions of these four metrics as features showing whether Alexa was able to provide these metrics for the target website.
These are given in rows 8--11 of Table~\ref{table:nonlin}.



\subsection{Audience Overlap Graph Construction}

When queried with a target news site's address, the Alexa {\em siteinfo}\footnote{http://www.alexa.com/siteinfo} tool returns a list of 4-5 sites that are most similar to the queried website based on audience overlap. For example, for \texttt{wsj.com}, we obtain the following list of similar websites and similarity scores: \texttt{marketwatch.com} 39.4, \texttt{cnbc.com} 39.4, \texttt{bloomberg.com} 35.9, \texttt{reuters.com} 34.5. We use these pairs of websites and overlap scores to build the edges of our graph, as shown in Figure~\ref{fig:wsj}.
Given a set of websites, we repeatedly query for each website and we grow our graph by adding new nodes and edges.
The resulting graph, obtained after performing this task for every site in our dataset, is referred to as level 0 audience overlap graph. 

\begin{figure}[htb]
\centering
\includegraphics[width=0.4\textwidth]{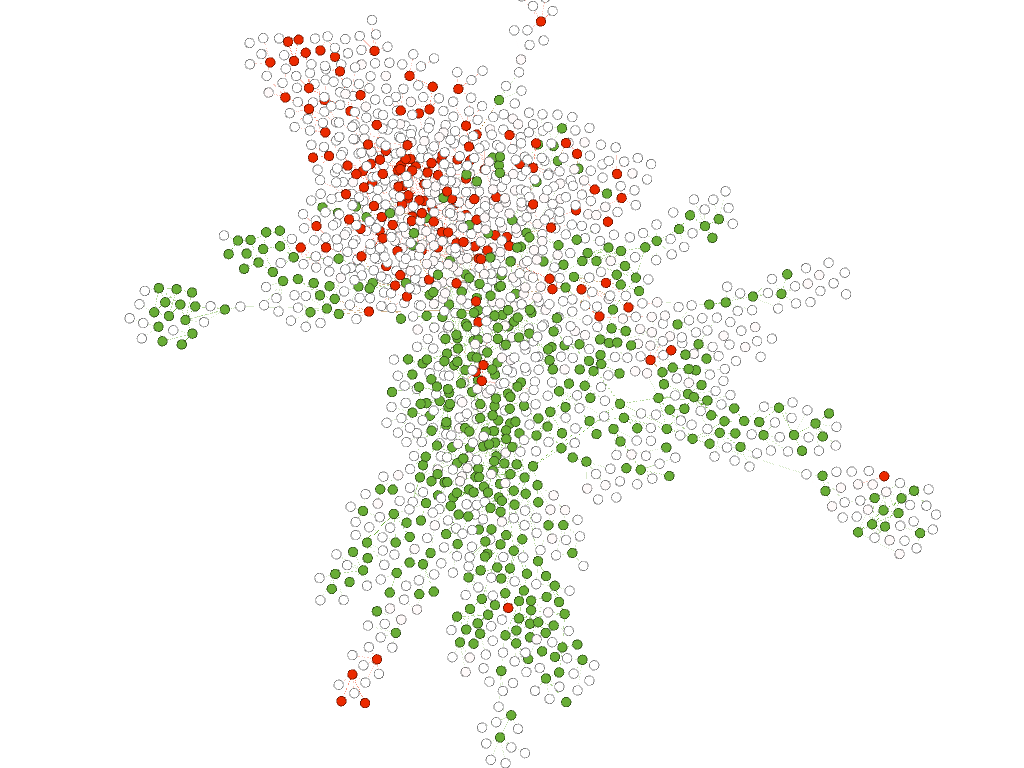}
\caption{Bird's eye view of our overlap graph. Nodes represent news sites and colors code site factuality: red corresponds to low-factuality, green to high-factuality, and white to mixed factuality and unknown sites.}
\label{fig:fact_bird_eye}
\end{figure}

For richer and denser representations, we then expand our overlap graph to higher levels. 
For this, we repeat the aforementioned steps of connecting website nodes according to audience overlap for the new websites identified during building the level-0 overlap graph, which were not initially in our seed list of websites. 
This yields to level-1 overlap graph as displayed in Figure~\ref{fig:fact_bird_eye}, where the distinction between low-factuality and high-factuality nodes can be clearly observed.
The same procedure is repeated until obtaining level-4 graphs. We observe that the performance gain is marginal beyond level-4. We attribute this observation to the weaker influence of the nodes towards the leaf to the labeled nodes as well as decreasing popularity of domains associated with the leaf nodes. Thus, we limit the expansion to level-4. 

\subsection{Representation Learning on Graphs}

In recent years, graph learning algorithms have been extensively used to model dependencies and relations between entities and to learn representations that embed graph nodes in a low-dimensional embedding space. We observe that different graph learning models learn different aspects of nodes in a graph. Thus, to get representations for news media in our overlap graphs, we use both random-walk based shallow graph embedding methods, such as node2vec \cite{grover2016node2vec}, and graph neural networks (GNNs) such as Graph Convolutional Networks (GCN)~\cite{gcn} and GraphSAGE~\cite{graphsage}.

In essence, Node2Vec \cite{grover2016node2vec} is one of the earliest graph learning frameworks. The model is inspired by Word2Vec \cite{mikolov2013efficient}, but instead of using sequences of words and optimizing the proximity loss, sequences for graph are generated by sampling random walks of a fixed maximum length for each node. 
These sequences of random walks are then used with a skip-gram model, just as with Word2Vec, to learn representations for the nodes. While Node2Vec produces embeddings solely based on the graph structure, GNN models, GCN and GraphSAGE, capture both the structure as well as their node/edge attributes. The latter models perform graph convolution operations over the computation graph of each node in the graph. A key difference between GCN and GraphSAGE is how they perform the convolution operation: GCN utilizes spectral operations whereas GraphSAGE utilizes spatial operations. Further, GCN considers all neighboring nodes whereas GraphSAGE is flexible to consider only a sampled subset of neighboring nodes. These differences in their constructions result in slightly different representations for different graph learning algorithms.

Using these three graph representation learning algorithms, we obtain low-dimensional vector representations (512 for Node2Vec, 128 for GCN, and 128 for GraphSAGE) of each node (website) in our graph. We empirically find that these embedding dimensions for respective algorithms yield the best downstream classification performance with a reasonable amount of computing resources.
We will refer to these representations as \emph{graph embeddings} throughout the paper.
When creating our models, 80\% of each dataset is assigned to the training set and 20\% to the test set, and 
for model evaluation we performed five-fold random cross validation.
Further, we tune our GNN models to use the following hyperparameters: number of epochs = 1000, number of layers = 4, learning rate = 0.01, weight decay = 0.0005 and dropout = 0.5. We tune our Node2Vec model to use the following hyperparameters: number of walks = 10, walk length = 100, number of dimensions = 512, return parameter (p) = 0.5 and in-out parameter (q) = 2.

\begin{table}[tbh]\centering
\resizebox{8cm}{!}{%
\begin{tabular}{lrlr|lrlrrr}\toprule
\multicolumn{4}{c}{\bf EMNLP-2018} &\multicolumn{4}{c}{\bf ACL-2020} \\\cmidrule{1-8}
\multicolumn{2}{c}{Political Bias} &\multicolumn{2}{c}{Factuality} &\multicolumn{2}{c}{Political Bias} &\multicolumn{2}{c}{Factuality} \\\cmidrule{1-8}
Left & 189 &High &256 &Left &243 &High &162 \\
Centre & 564 &Mixed &268 &Centre &272 &Mixed &249 \\
Right & 313 &Low &542 &Right &349 &Low &453 \\
\bottomrule
\end{tabular}}
\caption{Label distribution for the two datasets.}
\label{table:distdata}
\end{table}

\section{Experiments and Evaluation}
\label{sec:experiments}


\paragraph{Datasets}

To evaluate our system, we use two datasets from previous work: \cite{Ramy2018} and \cite{baly2020written}. We will refer to them as \emph{EMNLP-2018 dataset} and \emph{ACL-2020 dataset}, respectively. Both datasets contain lists of media domains along with their bias and factuality labels from Media Bias/Fact Check,\footnote{http://mediabiasfactcheck.com/} which is an independent journalism outlet. Factuality is modeled on a three-point scale, {\em i.e.},~\emph{high}, \emph{mixed}, and \emph{low}. Originally, political bias was modeled on a seven-point scale, but previous work has merged the fringe labels together and converted it into a three-point scale, {\em i.e.},~\emph{left}, \emph{centre}, and \emph{right}. Table~\ref{table:distdata} shows the label distribution of the two datasets.

\begin{table*}[tbh]
\centering 
\small

\begin{tabular}{l l c c} 
\toprule 
\bf \# & \bf Model & \bf F1 & \bf Acc. \\ [0.5ex] 
\midrule 
1 & Majority class baseline & 22.47 & 50.84 \\ 
\midrule
& \bf Previous work: \cite{Ramy2018} \\
2 & Articles (GloVe) & 58.02 & 64.35  \\
3 & Best overall model (Articles + Twitter + Wikipedia + URL analysis + Alexa Rank) & \bf \emph{59.91} & \bf \emph{65.48} \\
\midrule
& \bf Our results\\
4 & Node2Vec   & 60.60 & 68.19 \\ 
5 & GCN  & 72.23 & 75.94 \\ 
6 & Supervised GraphSage  & 86.04 & 87.55 \\ 
7 & Node2Vec+ Supervised GraphSage + GCN (late fusion) & 86.97 & 88.49 \\
8 & Node2Vec + Supervised GraphSage + GCN + Articles + AlexaMetrics (late fusion) & \bf 87.20 & \bf 88.58 \\
\bottomrule 
\end{tabular}
\caption{Factuality prediction on the EMNLP-2018 dataset.} 
\label{table:factuality18} 
\end{table*}

\hspace{0.5cm}

\begin{table*}[tbh]
\centering 
\small
\begin{tabular}{l l c c} 
\toprule 
\bf \# & \bf Model & \bf F1 & \bf Acc. \\ [0.5ex] 
\midrule 
1 & Majority class baseline & 22.93 & 52.43 \\
\midrule
 & \bf Previous work: \cite{baly2020written} \\
2 & Best ``Who Read It'' model & 42.48 & 58.76 \\
3 & Articles (BERT) & 61.46 & 67.94\\
4 & Best overall model (Articles + Twitter + YouTube) & \bf \emph{67.25} & \bf \emph{71.52} \\
\midrule
 & \bf Our results\\
5 & Node2Vec & 59.70 & 67.20 \\ 
6 & GCN & 53.76 & 61.47 \\ 
7 & Supervised GraphSage & 56.22 & 63.45 \\ 
8 & Node2Vec + Supervised GraphSage + GCN (late fusion) & 63.48 & 69.27 \\
9 & Node2Vec + Supervised GraphSage + GCN + Articles + Twitter + YouTube +  & \bf 69.61 & \bf 74.27 \\
& Facebook + AlexaMetrics (late fusion) &\\
\bottomrule 
\end{tabular}
\caption{Factuality prediction on ACL-2020 dataset.}
\label{table:factuality20} 
\end{table*}

\hspace{0.5cm}

\begin{table*}[tbh]
\centering 
\small
\begin{tabular}{l l c c} 
\toprule 
\bf \# & \bf Model & \bf F1 & \bf Acc. \\ [0.5ex] 
\midrule 
1 & Majority class baseline & 22.61 & 51.33 \\ 
\midrule
& \bf Previous work: \cite{Ramy2018}\\
2 & Articles (GloVe; our rerun) & 61.64 & 68.01 \\
3 & Best overall model (Articles + Wikipedia + URL analysis + Alexa Rank) & \bf \emph{63.27} & \bf \emph{69.89} \\ 
\midrule
 & \bf Our results\\
4 & Node2Vec & 67.64 &  73.55 \\
5 & GCN & 52.62 &  60.28 \\
6 & Supervised GraphSage & 52.18 &  64.81 \\
7 & Node2Vec + Supervised GraphSage + GCN (late fusion) & 65.97 &  73.20 \\
8  & Node2Vec + GCN + Supervised GraphSage + Articles + AlexaMetrics (late fusion) & \bf 72.44 & \bf 76.98 \\
\bottomrule 
\end{tabular}

\caption{Bias prediction on EMNLP-2018 dataset.} 
\label{table:bias18} 
\end{table*}

\begin{table*}[tbh]
\centering 
\small
\begin{tabular}{l l c c} 
    \toprule 
    \bf \# & \bf Model & \bf F1 & \bf Acc. \\ [0.5ex] 
    \midrule 
    1 & Majority Class & 19.18 &  40.39 \\ 
    \midrule
    & \bf Previous work: \cite{baly2020written}\\
    2 & Articles (BERT) & 79.34 & 79.75 \\
    3 & Best ``Who Read it'' model & 65.12 & 66.44 \\
    4 & Best overall model (Articles + Wikipedia + Twitter + YouTube) & \bf \emph{84.77} & \bf \emph{85.29} \\
    \midrule
    & \bf Our results\\
    5 & Node2Vec & 75.70 & 76.95 \\ 
    6 & GCN & 77.81 & 78.81 \\  
    7 & Supervised GraphSage & 88.50 & 88.59 \\ 
    8 & Node2Vec + GCN + Supervised GraphSage (late fusion) & 89.59 & 89.76 \\ 
    9 & Node2Vec + GCN + Supervised GraphSage + Articles + Wikipedia + Twitter + YouTube +  & \bf 91.93 & \bf 92.08\\
    &AlexaMetrics (late fusion)&\\
\bottomrule 
\end{tabular}
\caption{Bias Prediction on ACL-2020 dataset.}
\label{table:bias20} 
\vspace{-10pt}
\end{table*}

\vspace{-30pt}
\paragraph{Generation of Graph Embeddings} We use the audience overlap graph constructed above along with the all the Alexa site metrics as node features. 
We then impute the missing features by taking the average of the five nearest neighbors. Both GCN and GraphSAGE are executed under the semi-supervised setting and the representations of the last hidden layer in the respective models are obtained as the node embeddings.  
Node2vec takes the graph structure as the input and produces a node embedding for each node in an unsupervised setting.



\paragraph{Experimental Setup}
The predictive capability of the node-level representations obtained using the three graph learning models are evaluated both individually and in combination in a supervised setting.
For this, we used five-fold cross-validation to train and to evaluate an SVM model using the the node embeddings along with their factuality and bias labels. We performed grid search to tune the values of the hyper-parameters of our SVM model with an RBF kernel.
As the datasets for both years and for both tasks are imbalanced, we optimized macro-F1 using grid search. We evaluated our model on the remaining unseen fold, and we report both macro-F1 score and accuracy.

When combining the three representations, we adopt a late-fusion strategy.
To this end, we train separate classifiers for each type of representation, and then we train an ensemble by averaging the posterior probabilities obtained by each model. 
This allows the ensemble model to learn different weights, thereby ensuring that more  
attention is paid to the probabilities produced by better models. 

Finally, to evaluate the complementary nature of the audience homophily and 
characteristics exhibited in the textural content of media websites and descriptions of their audience in social media platforms, these two sources of information are combined to make predictions.
For this purpose, we utilized the sentence representations of the news articles and Wikipedia descriptions associated with each news medium as well as the Twitter, YouTube, and Facebook audience representations available in the repository of \cite{baly2020written}. 
Further details on these features are provided in Sec.\ref{sec:appB} of the Appendix.

For studying the efficacy of our system, we compare the results of EMNLP-2018 dataset to the best previous overall models and with models using only textual representations (which was the best-performing single feature and included GloVe~\cite{pennington-etal-2014-glove} representations for the articles). 
As our audience overlap graph falls under the \emph{Who Read It} category of features in \cite{baly2020written}, for the 2020 tasks, in addition to the best previous model and the best model using textual representations (based on average RoBERTa sentence representations), we also compare to the best \emph{Who Read It} model.

We used Nvidia's K80 GPUs to train the graph embeddings, which took around 30 minutes.
The neural network training and inference phases were both carried out on the CPU.
In our repository we've documented every package version for easy replication of our results. 

\paragraph{Factuality Prediction}
Table~\ref{table:factuality18} shows our results for factuality prediction task on the EMNLP-2018 dataset. 
(In the table, each group of embeddings are referred to by the name of graph learning algorithm used in their generation.)
We can see that all three types of graph embeddings (rows 4-6) outperform the Articles representations (row 2) and the best result from previous work (row 3), which combines representations from several sources. 
As expected, the combination of the graph embeddings perform as a more powerful predictor improving the macro-F1 score by more than 17 points absolute (row 7).
We then incorporated our graph representations with a subset of earlier introduced features that yielded the best performance (row 8). 
This provided an additional improvement of +0.23  macro-F1 points.

Table~\ref{table:factuality20} shows our results on the ACL-2020 dataset for the factuality prediction task. Here our graph embeddings (rows 5-7) perform comparable to the best text representation from previous work (row 3), i.e., the Articles representation obtained using fine-tuned BERT. 
Comparing the graph embeddings with other audience characteristics (the \emph{Who Read It} category of features), we can see that the discrimination power inherent to the audience overlap feature is much higher (by around 14-17 macro-F1 points absolute) than that of the latter features.
Unlike the EMNLP-2018 dataset, however, none of the graph embeddings outperformed the best model that combines different versions of textual representations (row 4).
Further, we observed that Node2Vec representations yielded more accurate predictions than GCN and GraphSAGE representations on this dataset. 
When graph embeddings are combined with earlier introduced representations, those associated with Articles and descriptive characteristics of Twitter, YouTube, and Facebook audiences, 
we obtain a significantly better result that outperforms the previous best result by a margin of +2.36 macro-F1 score (row 9).
This result also confirms that graph embeddings are complementary to the textual representations. 

\paragraph{Bias Prediction}

Table~\ref{table:bias18} shows evaluation results for bias detection on the EMNLP-2018 dataset.
Here, we observe that among the three graph embeddings (rows 4-7), only Node2Vec embeddings outperformed the previous best overall model (row 3). 
The ensemble classifier's accuracy (row 7) was expectedly very similar to that of the top performing classifier.
Although the graphs embeddings could not produce superior performance, their combination with textual and other audience features yielded a substantial increase of +9.17 macro-F1 points absolute over the best previous result (row 8).
This further confirms the complementarity of audience homophily and textual representations on the bias detection task.

Table~\ref{table:bias20} shows the corresponding results for the ACL-2020 dataset. 
For this setting GraphSage embeddings (row 7) are determined to produce significantly more accurate predictions than both other embeddings (rows 5-6) and the previous best overall model (row 4). 
The ensemble system was also able to leverage the strengths of the three types of graph embeddings and yielded the best performance (row 8).
When the graph embeddings capturing the audience homophily characteristics are combined with other representations (row 9), the improvement in performance is further enhanced by an overall increase of +7.16 macro-F1 score points over the previous best overall result.


\section{Discussion}
\label{sec:discussion}




\paragraph{Other Features Tested} 
    Alexa Site Info maintains a wide array of audience centric statistics for the websites. Apart from audience overlap, we also experimented with other features: \emph{Alexa Rank}, \emph{Total Sites Linking In}, \emph{Daily Page Views per Visitor}, \emph{Bounce Rate}, \emph{Average Daily Time per Visitor}. Table~\ref{table:nonlin} shows that these features performed better than the majority class baselines,  they are not very strong. Note that most of these features were heavily unpopulated for a substantial part of our website dataset, which could be the reason for their mediocre performance. 
    Regardless, site popularity and engagement metrics are potentially very useful for bias and factuality prediction.
    In fact, as our results show, even their binarized versions are helpful, even on top a very a strong system. 

\paragraph{Varying Predictive Power of Graph Learning Methods}
Our results show that none of the three graph learning approaches perform consistently better than the others. 
Most surprisingly, we determined that Node2Vec algorithm yielded better predictions than the two GNN models, GCN and GraphSAGE, in two settings. We believe an important factor contributing to this result is the sparsity of node features.
As can be seen in Table \ref{table:nonlin}, among all news media websites comprising our auidence overlap graph, just three Alexa metrics were available for more than 94\% of the websites.
Whereas four metrics were available only for less than 40\% of the websites.
Since GNNs' superior performance primarily stem from their ability to combine graph structure and node information, the missing features likely curtailed their performance significantly. 
In fact, our earlier tests performed without imputing missing features yielded a much lower accuracy results.
Therefore, it is plausible to assume that the performance of GNN models will increase in the presence of more discriminant node features.

Further, our analysis revealed that the features {\em Sites Linking In}, {\em Alexa Rank} and {\em Daily Time on Site} are more important than the other two features {\em Bounce Rate} and {\em Daily Pageviews per Visitor} for both tasks of factuality and bais prediction. However, there is a slight variation in the order of importance for these tasks. For example, {\em Alexa Rank} was the most important feature for factuality prediction whereas {\em Sites Linking In} was the most important feature for bias prediction. Combining the features together with the graph structure assisted in improving the performance of both tasks.


\paragraph{Different Levels} 
    Our preliminary experiments have shown that, as we use embeddings from higher level graphs, performance improves. Table~\ref{table:embeddingslevels} shows our results on incremental levels of graphs on the EMNLP-2018 factuality dataset. We can notice a jump of +15.40 macro-F1 points absolute when going from a level-0 to a level-4 graph. This increase in performance can be attributed to the addition of more nodes and denser connections between them in the graph, which enhances our graph embeddings. Based on these preliminary results, we decided to use level 4 embeddings as our overlap graph embeddings in all our experiments.
    
\paragraph{\emph{Who Read It} vs. \emph{What Was Written} Features} 
     With the introduction of graph embeddings in the \emph{Who Read It} feature category, we narrowed the gap between \emph{What Was written} and \emph{Who Read It} features, as reported in \cite{baly2020written}.

\begin{table}[ht]
\centering 
\small
\begin{tabular}{@{}llccc@{}} 
\toprule 
\bf \# & \bf Model & \bf \% Pop. &  \bf F1 & \bf Acc.  \\ [0.5ex] 
\midrule 
1 & Majority class baseline  & -- & 22.47 & 50.84 \\ 
2 & Alexa Rank (reciprocal) & 99.92 & 22.46 & 50.75 \\
\midrule
3 & Alexa Rank (logarithm) & 99.92 & 44.81 & 55.07 \\
4 & Total Sites Linking In & 94.98 & 45.28 & 55.72  \\
5 & Bounce Rate & 31.09 & 44.70 & 55.25  \\
6 & Average Daily Time & 36.27 & 44.13 & 56.10  \\
7 & Daily Pageviews & 61.08 & 44.93 & 56.85  \\
\midrule
8 & Has Total Sites Linking In & 94.98 & 23.03 & 50.94  \\
9 & Has Bounce Rate & 31.09 & 42.70 & 59.38  \\
10 & Has Average Daily Time & 36.27 & 42.50 & 59.47  \\
11 & Has Daily Pageviews & 61.08 & 37.19 & 56.10  \\

\midrule
12 & Combination of 3--7 & -- & 48.14 & 57.50\\
\midrule
13 & Combination of 8--11 & -- & 43.08 & 59.19\\
\bottomrule 
\end{tabular}
\caption{Factuality prediction on the EMNLP-2018 dataset using different statistics from Alexa. Line 2 shows a result from \cite{Ramy2018}. Line 12 combines lines 3--7, and line 13 combines lines 8--11. For missing values, we take the mean value of the feature.} 
\label{table:nonlin} 
\end{table}

\begin{table}[ht]
\centering 
\small
\begin{tabular}{l r r c c } 
\toprule 
\bf Model & \bf Nodes & \bf Edges & \bf F1 & \bf Acc. \\ 
\midrule 
Majority & -- & -- & 22.47 & 50.84 \\ 
\midrule
level 0 & 1,062 & 4,837 & 45.20  & 57.50\\
level 1 & 4,238 & 20,335 & 55.80 & 64.70\\ 	
level 2 & 11,867 & 57,320 & 56.78 & 65.01\\
level 3 & 30,889 & 149,110 &  57.70 & 66.10\\
level 4 & 78,429 & 377,260 & 60.60 & 68.19\\
\bottomrule 
\end{tabular}
\caption{Ablation study: factuality prediction on the EMNLP-2018 data using Node2Vec graph embeddings from graphs of different levels of expansion.} 
\label{table:embeddingslevels} 
\end{table}
\vspace{-20pt}



\paragraph{Alternatives to Alexa Siteinfo}
Alexa siteinfo service was discontinued in May 2022. While our approach relies on Alexa siteinfo to obtain the audience overlapping information and Alexa matrices for domains, we believe that our approach is generic in that one may utilize an alternative SEO data source such as Ahref, Semrush, Similarweb or Moz to obtain similar information to construct the audience overlap graph and extract features for the websites under consideration. We leave it as a future direction to explore these alternative sources. Further different SEO sources have different coverage of websites, one may combine multiple such sources to not only address the missing features but also to increase the number of websites our approach connects leading to the discovery of additional biased or low factual websites. 

\section{Conclusion and Future Work}
\label{sec:conclusion}

We studied the problem of media profiling with respect to their factuality of reporting and bias. Motivated by homophily considerations, 
we built a graph of inter-media connections based on the audience overlap for the target pair of news media, and then we used graph neural networks to come up with representations for each medium. We found that such representations, especially when augmented with Alexa Metrics and additional information sources from Twitter, Facebook, YouTube, and Wikipedia, are quite useful, yielding state-of-the-art results on four standard datasets for predicting the factuality and the bias of news media.

In future work, we plan to experiment with other kinds of graph neural networks. We further want to integrate additional information sources.

\section*{Limitations}
Our work relies on the Alexa website ranking and traffic information to build the input graphs, but Alexa is to be discontinued in the future. However, we envision that one may use similar alternative tools such as Semrush, Ahrefs or SimilarWeb to build similar graphs and use the proposed approach. 

Our work excludes isolated nodes (websites) in the constructed graph. Such isolated nodes could occur when a website is either relative new or not profiled in Alexa due to insufficient traffic. This in turn results in some websites not being able to be classified. The datasets used in this work have only one such isolated website, but we suggest utilizing non-graph information, as used in prior approaches, to classify them.


\section*{Ethics and Broader Impact}

\paragraph{Data Collection}
We collected the data for our graph using the Alexa Audience Overlap Tool.\footnote{\url{http://alexa.com/marketing-stack/audience-overlap-tool}}
Although obtained Alexa statistics provide an extensive view of audience overlap across media sites, it is not comprehensive as they are only limited to top-five sites for each query. 
Further, sites with fewer audience are likely to be more prone to measurement error, therefore inferring factuality and bias ratings of those sites is more challenging.

\paragraph{Biases}
There might be biases in our gold labels from Media Bias/Fact Check, as in some judgments for factuality and bias might be subjective. These biases, in turn, will likely be exacerbated by the supervised models trained on them \cite{BIAS}. This is beyond our control, as are the potential biases in pre-trained large-scale transformers such as BERT and RoBERTa, which we use in our experiments.

\paragraph{Intended Use and Potential Misuse}
Our models can enable analysis of entire news outlets, which could be of interest to fact-checkers, journalists, social media platforms, and policymakers.
Yet, they could also be misused for malicious attacks like targeting specific parts of the audience with misinformation news. We, therefore, ask researchers to exercise caution.

\paragraph{Environmental Impact}
We would also like to warn that the use of large-scale Transformers requires a lot of computations and the use of GPUs/TPUs for training, which contributes to global warming \cite{strubell-etal-2019-energy}. This is a bit less of an issue in our case, as we do not train such models from scratch; rather, we fine-tune them on relatively small datasets. Moreover, running on a CPU for inference, once the model is fine-tuned, is perfectly feasible, and CPUs contribute much less to global warming.

\bibliographystyle{acl_natbib}
\bibliography{custom}

\newpage
\clearpage
\section*{Appendix}
\label{sec:appendix}
\appendix
\section{Alexa Audience Overlap Scores}
\label{sec:appA}

\begin{figure}[h]
\centering
\includegraphics[scale=0.30]{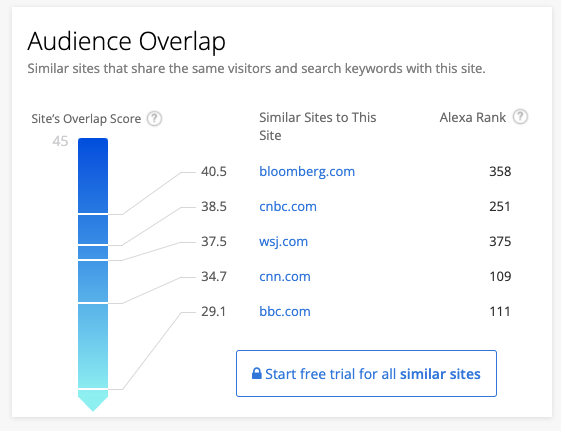}
\caption{Alexa Rank audience overlap for \emph{reuters.com}.}
\label{fig:alexa-audience-overlap-reuters}
\end{figure}

\begin{figure}[h]
\centering
\includegraphics[scale=0.30]{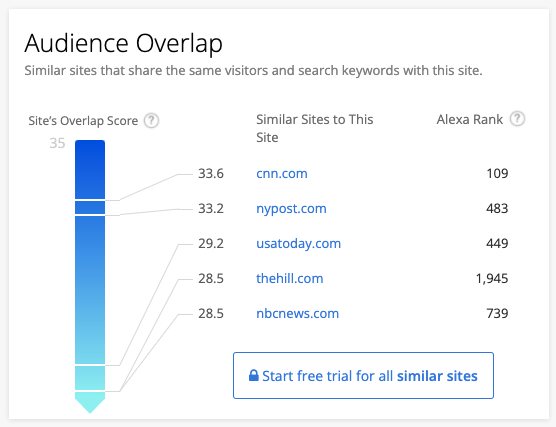}
\caption{Alexa Rank audience overlap for \emph{foxnews.com}.}
\label{fig:alexa-audience-overlap-foxnews}
\end{figure}

\begin{figure}[h]
\centering
\includegraphics[scale=0.30]{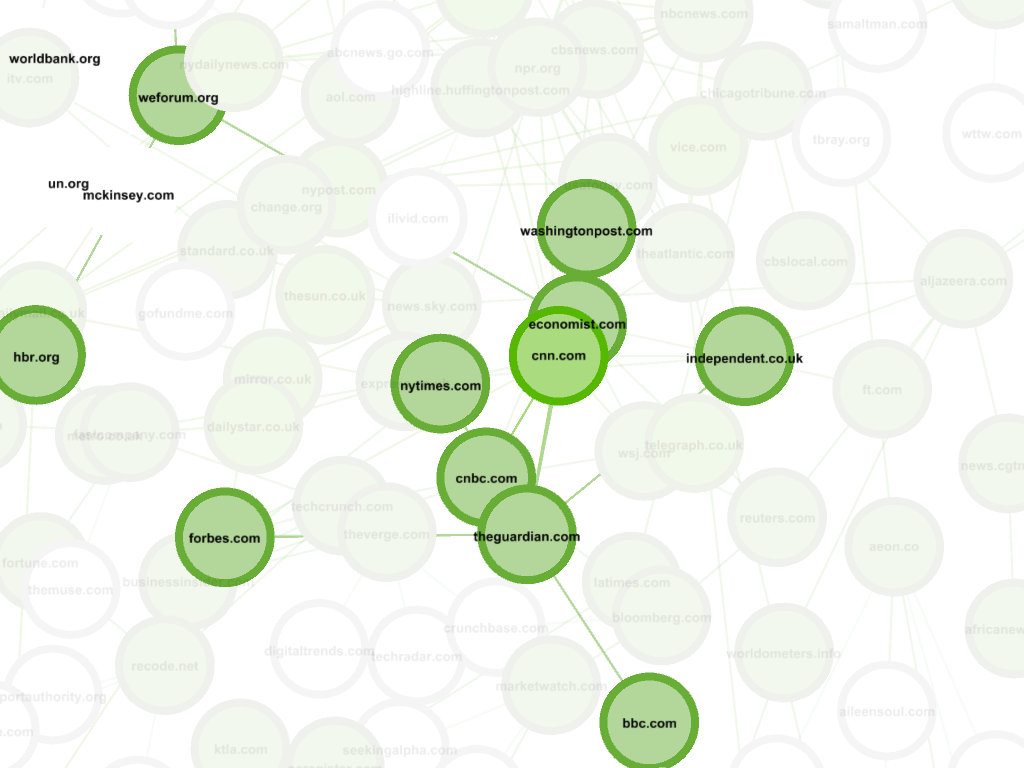}
\caption{Alexa Rank audience overlap for \emph{cnn.com}.}
\label{fig:alexa-audience-overlap-cnn}
\end{figure}

\begin{figure}[h]
\centering
\includegraphics[scale=0.30]{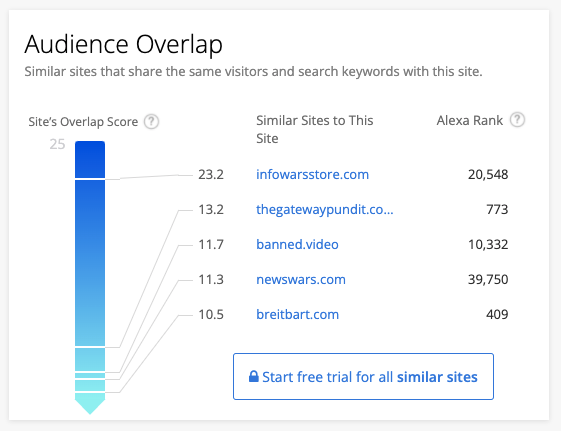}
\caption{Alexa Rank audience overlap for \emph{infowars.com}.}
\label{fig:alexa-audience-overlap-infowars}
\end{figure}

Figures~\ref{fig:alexa-audience-overlap-reuters}--\ref{fig:alexa-audience-overlap-infowars} show the information provided by Alexa's Audience Overlap tool for \emph{reuters.com}, \emph{foxnews.com}, \emph{cnn.com}, and \emph{infowars.com}. We can see that a highly factual site, such as reuters.com, has sizable audience overlap with other factual sites. Similarly, a low-factuality website such as \emph{infowars.com}, shares audience with  other low-factuality websites. The audience homophily also holds for political bias as can be seen in cases of \emph{foxnews.com} and \emph{cnn.com}.

\section{Supplementary Data Sources}
\label{sec:appB}
In addition to characterizing audience homophily, we also consider textual information sources about news media available on the Web and the audience characteristics of their social media platforms. These include the following.

{\em News Articles and Wikipedia:}
Previous work on the task used either GloVe \cite{Ramy2018} or fine-tuned BERT encodings \cite{baly2020written} of the news articles, and averaged these encodings across articles by the website to obtain a textual representation for the website/domain. Similarly, GloVe and pre-trained BERT were used to get encodings for the Wikipedia descriptions of media. 
Thus, we also used articles and Wikipedia descriptions to obtain site-level textual representations. 
For the EMNLP-2018 Bias and Factuality tasks, 
we used the averaged GloVe encodings of the articles present on the website. For the ACL-2020 Bias and Factuality tasks,
we used sentence encoders based on RoBERTa \cite{reimers2019sentence} to encode the text, \emph{i.e.},~the articles or Wikipedia descriptions. For news media without a Wikipedia page, we used a vector of zeroes. We refer to these textual representations as \emph{Articles} and \emph{Wikipedia}. 


{\em Audience Characteristics:}
To model the similarity between news media in terms of the overlap of their audience and of quantifying the level of engagement between a medium and its followers, we also obtained an audience-centric representation for each medium, by considering the users of social media platforms that have interest in the content created by these news sources. 
For this purpose, we considered three features that were reported to perform well in characterization of followers of a news medium  \cite{baly2020written}.

The first feature is based on how Twitter users following the account of the medium self-describe in their publicly accessible Twitter profiles.  
For each medium, this is obtained by encoding the biographic descriptions of 5,000 English-speaking Twitter followers, using BERT and obtaining an average representation.
The second feature involves how audience of the medium’s YouTube channel respond to each video in terms of the number of comments, views, likes and dislikes; by averaging these statistics over all videos, another  medium-level representation is generated. 
The last feature includes audience estimates from Facebook's advertising platform which is used to obtain demographic information for the audience interested in each medium; this data is used to obtain the audience distribution over the political spectrum, the distribution is then divided into five categories, and each medium is labeled accordingly. 
These three features are referred to as \emph{Twitter}, \emph{YouTube}, and \emph{Facebook} audience representations. 

\end{document}